\documentclass[letterpaper]{article}
\usepackage{aaai18}
\usepackage{times}
\usepackage{helvet}
\usepackage{courier}
\usepackage{url}
\usepackage{graphicx}
\usepackage{amsfonts}
\usepackage{dsfont}
\usepackage{amsmath}
\usepackage{amssymb}
\usepackage{diagbox}
\usepackage[table,xcdraw]{xcolor}
\usepackage[T1]{fontenc}

\def\imsize{.4cm}

\title{Deep Learning for Case-Based Reasoning through Prototypes:\\ A Neural Network that Explains Its Predictions}

\author{Oscar Li\thanks{\protect Contributed equally}\textsuperscript{1}, Hao Liu\footnotemark[1]\textsuperscript{3}, Chaofan Chen\textsuperscript{1}, Cynthia Rudin\textsuperscript{1,2}\\
\textsuperscript{1}Department of Computer Science, Duke University, Durham, NC, USA 27708\\
\textsuperscript{2}Department of Electrical and Computer Engineering, Duke University, Durham, NC, USA 27708\\
\textsuperscript{3}Kuang Yaming Honors School, Nanjing University, Nanjing, China, 210000\\
runliang.li@duke.edu, 141242059@smail.nju.edu.cn, 
\{cfchen, cynthia\}@cs.duke.edu}
\begin{document}
\nocopyright
\maketitle
\begin{abstract}
Deep neural networks are widely used for classification. These deep models often suffer from a lack of interpretability -- they are particularly difficult to understand because of their non-linear nature. As a result, neural networks are often treated as ``black box'' models, and in the past, have been trained purely to optimize the accuracy of predictions. In this work, we create a novel network architecture for deep learning that naturally explains its own reasoning for each prediction. This architecture contains an autoencoder and a special {\it prototype layer}, where each unit of that layer stores a weight vector that resembles an encoded training input. The encoder of the autoencoder allows us to do comparisons within the latent space, while the decoder allows us to visualize the learned prototypes. The training objective has four terms: an accuracy term, a term that encourages every prototype to be similar to at least one encoded input,  a term that encourages every encoded input to be close to at least one prototype, and a term that encourages faithful reconstruction by the autoencoder. The distances computed in the prototype layer are used as part of the classification process. Since the prototypes are learned during training, the learned network naturally comes with explanations for each prediction, and the explanations are loyal to what the network actually computes.
\end{abstract}

\noindent 


\section{Introduction}
As machine learning algorithms have gained importance for important societal questions, interpretability (transparency) has become a key issue for whether we can trust predictions coming from these models. There have been cases  where incorrect data fed into black box models have gone unnoticed, leading to unfairly long prison sentences (e.g., prisoner Glen Rodriguez was denied parole due to an incorrect COMPAS score, \citeauthor{nyt-computers-crim-justice}, \citeyear{nyt-computers-crim-justice}). In radiology, lack of transparency causes challenges to FDA approval for deep learning products. Because of these issues, ``opening the black box'' of neural networks has become a debated issue in the media \cite{forbesunfairness,Smith2016,propublica2016,Westervelt2017}. Artificial neural networks are particularly difficult to understand because their highly nonlinear functions do not naturally lend to an explanation that humans are able to process.

In this work, we create an architecture for deep learning that explains its own reasoning process. The learned models naturally come with explanations for each prediction, and the explanations are loyal to what the network actually computes. As we will discuss shortly, creating the architecture to encode its own explanations is in contrast with creating explanations for previously trained black box models, and aligns more closely with work on prototype classification and case-based reasoning. 


In the past, neural networks have often been designed purely for accuracy, with \textit{posthoc} interpretability analysis. In this case, the network architecture was chosen first, and afterwards one aims to interpret the trained model or the learned high-level features. To do the interpretability analysis requires a separate modeling effort. One problem with generating explanations posthoc is that the explanations themselves can change based on the model for the explanation. For instance, it may be easy to create multiple conflicting yet convincing explanations for how the network would classify a single object, none of which are the correct reason for why the object was classified that way.
A related issue is that posthoc methods often create explanations that do not make sense to humans. This means that extra modeling is needed to ensure that the explanations are interpretable. This happens, for instance, in the Activation Maximization (AM) approach, where one aims to find an input pattern that produces a maximum model response for a quantity of interest to the user \cite{Erhan}. Since the images from AM are not generally interpretable (they tend to be gray), regularized optimization is used to find an interpretable high activation image \cite{hinton2012practical,lee2009convolutional,Oord16,Nguyen16}. When we add regularization, however, the result is a combination of what the network actually computes and the extrinsic regularization.  Given that the explanations themselves come from a separate modeling process with strong priors that are not part of training, we then wonder how we can trust the explanations from the posthoc analysis. In fact there is a growing literature discussing the issues mentioned above for AM \cite{interpret_review}. For images, posthoc analysis often involves visualization of layers of a neural network. For instance, an alternative to AM was provided by \citeauthor{ZeilerFe14} \shortcite{ZeilerFe14}, who use deconvolution as a technique to visualize what a convolutional neural network (CNN) has learned. Deconvolution is one method for decoding; our method can use any type of decoder to visualize the prototypes, including deconvolution. In addition, \citeauthor{ZeilerFe14} \shortcite{ZeilerFe14} try to visualize parts of images that most strongly activate a given feature map, but they do not provide an explanation for how the network reaches its decision. In contrast, we build a reasoning process into our network and do not consider posthoc analysis in this work.

There are other works that also build interpretability into deep neural networks without using posthoc analysis. \citeauthor{pinheiro2015image} \shortcite{pinheiro2015image} design a network for weakly supervised image segmentation by training a classification network that extracts important pixels which could potentially belong to an object of some class. \citeauthor{Regina16_NLP} \shortcite{Regina16_NLP} propose a network architecture that extracts parts of an input as a rationale and uses the rationale for predictions. Both of these works build interpretability into neural networks by extracting parts of an input and focusing on those parts for their respective tasks. Our method differs in that we use case-based reasoning instead of extractive reasoning -- our model explains its predictions based on similarity to prototypical cases, rather than highlighting the most relevant parts of the input; it is possible for their ideas to be combined with ours. \citeauthor{tan2015improving} \shortcite{tan2015improving} and \citeauthor{wu2016stimulated} \shortcite{wu2016stimulated} 
aim to improve the interpretability of activation patterns of feature maps in deep neural networks used for speech recognition. In contrast, our model does not aim to enforce a particular structure on feature maps -- it allows flexibility in feature learning but introduces a special prototype layer for decision interpretation.


Our network is a form of \textit{prototype classifier}, where observations are classified based on their proximity to a prototype observation within the dataset. For instance, in our handwritten digit example, we can determine that an observation was classified as a ``3'' because the network thinks it looks like a particular prototypical ``3'' within the training set. If the prediction is uncertain, it would identify prototypes similar to the observation from different classes, e.g., ``4'' is often hard to distinguish from ``9'', so we would expect to see prototypes of classes 4 and 9 identified when the network is asked to classify an image of a 9.

Our work is closely aligned with other prototype classification techniques in machine learning \cite{Bien2011,KimRuSh14,Priebe2003,WuTabak2017}. Prototype classification is a classical form of case-based reasoning \cite{kolodner1992introduction}; however, because our work uses neural networks, the distance measure between prototypes and observations is measured in a flexible latent space. The fact that the latent space is adaptive is the driving force behind its high quality performance. 

The word ``prototype'' is overloaded and has various meanings. For us, a prototype is very close or identical to an observation in the training set, and the set of prototypes is representative of the whole data set. In other contexts, a prototype is not required to be close to any one of the training examples, and could be just a convex combination of several observations.
In few-shot and zero-shot learning, prototypes are points in the feature space used to represent a single class, and distance to the protoype determines how an observation is classified. 
For example, ProtoNets \cite{protonet} utilize the mean of several embedded ``support'' examples as the prototype for each class in few-shot learning. \citeauthor{zero_shot_prototype} \shortcite{zero_shot_prototype} use a generative probabilistic model to generate prototypes for zero shot learning, which are points in the feature space. 
In both cases, prototypes are not optimized to resemble actual observations, and are not required to be interpretable (meaning that their visualizations will not generally resemble natural images), and each class can have only one prototype. 

Our deep architecture uses an autoencoder \cite{autoencoder} to create a latent low-dimensional space, and distances to prototypes are computed in that latent space. 
Using a latent space for distance computation enables us to find a better dissimilarity measure than $L^2$ on the pixel space. 
Other works also use latent spaces, e.g., \citeauthor{SalakhutdinovH07} \shortcite{SalakhutdinovH07} conduct a soft $k$-nearest neighbors classification on the latent space of a restricted Boltzman machine autoencoder, although not for the aim of interpretability.



\section{Methodology}

\subsection{Network Architecture}
\begin{figure*}
\centering
\includegraphics[width=0.9\textwidth]{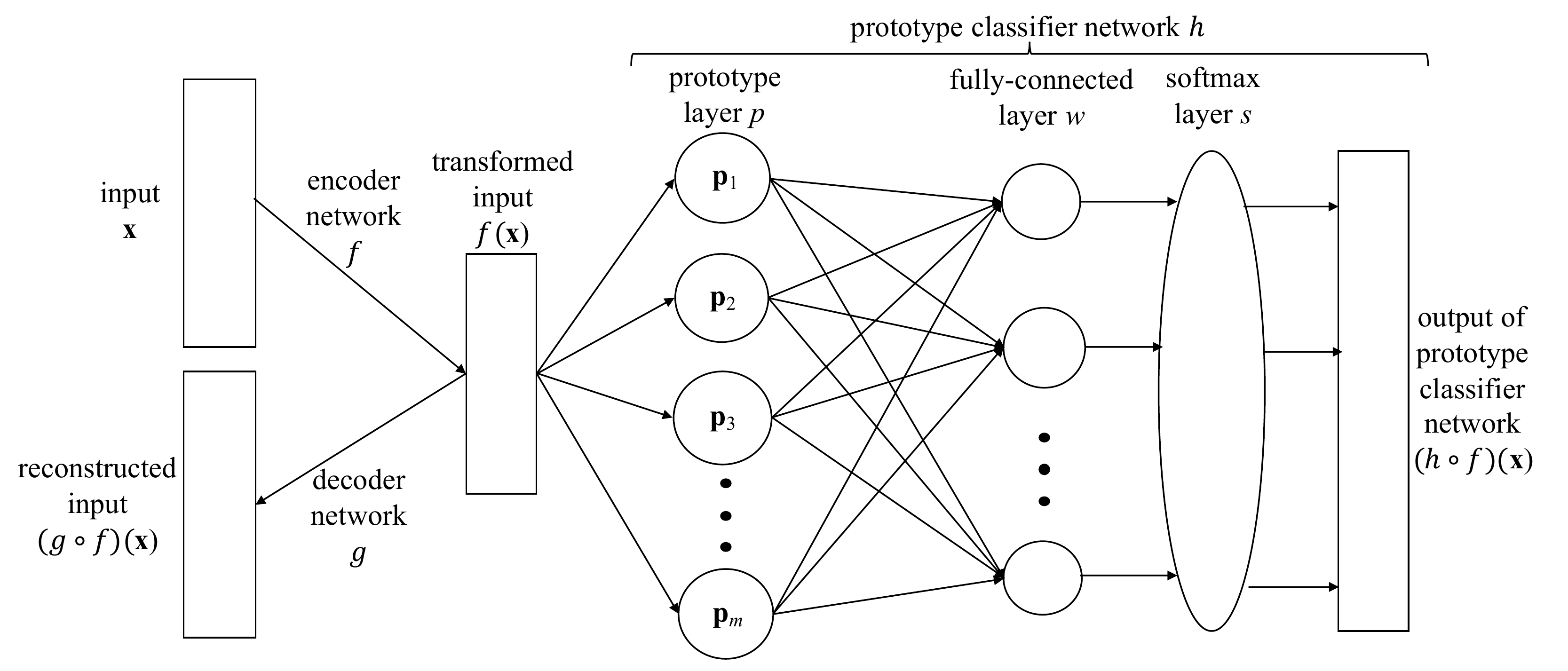}
	\caption{Network Architecture}
	\label{fig:network}
\end{figure*}
Let $D = \{(\mathbf{x}_i, y_i)\}_{i=1}^n$ be the training dataset with $\mathbf{x}_i \in \mathbb{R}^p$ and $y_i \in \{1, ..., K\}$ for each $i \in \{1, ..., n\}$. Our model architecture consists of two components: an autoencoder (including an encoder, $f: \mathbb{R}^p \rightarrow \mathbb{R}^q$, and a decoder, $g: \mathbb{R}^q \rightarrow \mathbb{R}^p$) and a prototype classification network $h: \mathbb{R}^q \rightarrow \mathbb{R}^K$, illustrated in Figure \ref{fig:network}. The network uses the autoencoder to reduce the dimensionality of the input and to learn useful features for prediction; then it uses the encoded input to produce a probability distribution over the $K$ classes through the prototype classification network $h$. The network $h$ is made up of three layers: a prototype layer, $p: \mathbb{R}^q \rightarrow \mathbb{R}^m$, a fully-connected layer $w: \mathbb{R}^m \rightarrow \mathbb{R}^K$, and a softmax layer, $s: \mathbb{R}^K \rightarrow \mathbb{R}^K$. The network learns $m$ prototype vectors $\mathbf{p}_1, ... , \mathbf{p}_m \in \mathbb{R}^q$ (each corresponds to a {\it prototype unit} in the architecture) in the latent space. The prototype layer $p$ computes the squared $L^2$ distance between the encoded input $\mathbf{z}=f(\mathbf{x}_i)$ and each of the prototype vectors:
\begin{equation}
\label{eq:prototype_layer}
p(\mathbf{z}) = \begin{bmatrix} \|\mathbf{z}- \mathbf{p}_1\|_2^2, & \|\mathbf{z}- \mathbf{p}_2\|_2^2, & ... & \|\mathbf{z}- \mathbf{p}_m\|_2^2\\  \end{bmatrix}^\top.
\end{equation}
In Figure \ref{fig:network}, the \textit{prototype unit} corresponding to $\mathbf{p}_j$ executes the computation $\|\mathbf{z}- \mathbf{p}_j\|_2^2$. The fully-connected layer $w$ computes weighted sums of these distances $Wp(\mathbf{z})$, where $W$ is a $K \times m$ weight matrix. These weighted sums are then normalized by the softmax layer $s$ to output a probability distribution over the $K$ classes. The $k$-th component of the output of the softmax layer $s$ is defined by
\begin{equation}
s(\mathbf{v})_k = \frac{\exp(v_k)}{\sum_{k'=1}^K \exp(v_{k'})}
\end{equation}
where $v_k$ is the $k$-th component of the vector $\mathbf{v}=Wp(\mathbf{z}) \in \mathbb{R}^K$. 

During prediction, the model outputs the class that it thinks is the most probable. In essence, our classification algorithm is distance-based on the low-dimensional learned feature space. A special case is when we use one prototype for every class (let $m = K$) and set the weight matrix of the fully-connected layer to the negative identity matrix, $W = -I_{K\times K}$ (i.e. $W$ is not learned during training). Then the data will be predicted to be in the same class as the nearest prototype in the latent space. More realistically, we typically do not know how many prototypes should be assigned to each class, and we may want a different number of prototypes from the number of classes, i.e., $m \neq K$. In this case, we allow $W$ to be learned by the network, and, as a result, the distances to all the prototype vectors will contribute to the probability prediction for each class.

This network architecture has at least three advantages. First, unlike traditional case-based learning methods, the new method automatically learns useful features. For image datasets, which have dimensions equal to the number of pixels,
if we perform classification
using the original input space or use hand-crafted feature spaces, the methods tend to perform poorly (e.g., $k$-nearest neighbors). 
Second, because the prototype vectors live in the same space as the encoded inputs, we can feed these vectors into the decoder and visualize the learned prototypes throughout the training process. This property, coupled with the case-based reasoning nature of the prototype classification network $h$, gives users the ability to interpret how the network reaches its predictions and visualize the prototype learning process without \textit{posthoc} analysis. Third, when we allow the weight matrix $W$ to be learnable, we are able to tell from the strengths of the learned weight connections which prototypes are more representative of which class. 
\subsection{Cost Function}
The network's cost function reflects the needs for both accuracy and interpretability. In addition to the classification error, there is a (standard) term that penalizes the reconstruction error of the autoencoder. There are two new error terms that encourage the learned prototype vectors to correspond to meaningful points in the input space; in our case studies, these points are realistic images. All four terms are described mathematically below.

We use the standard cross-entropy loss for penalizing the misclassification. The cross-entropy loss on the training data $D$ is denoted by $E$, and is given by
\begin{equation}\label{eq:cross-entropy}
E(h \circ f, D) = \frac{1}{n}\sum_{i=1}^n\sum_{k=1}^K - \mathds{1}[y_i = k]\log((h \circ f)_k(\mathbf{x}_i))
\end{equation}
where $(h \circ f)_k$ is the $k$-th component of $(h \circ f)$. We use the squared $L^2$ distance between the original and reconstructed input for penalizing the autoencoder's reconstruction error.  The reconstruction loss, denoted by $R$, on the training data $D$ is given by
\begin{equation}\label{eq:reconstr-loss}
R(g \circ f, D) = \frac{1}{n}\sum_{i=1}^n \|(g \circ f)(\mathbf{x}_i) - \mathbf{x}_i\|_2^2.
\end{equation}
The two interpretability regularization terms are formulated as follows:
\begin{equation}\label{eq:reg_1}
R_1(\mathbf{p}_1, ..., \mathbf{p}_m, D) = \frac{1}{m}\sum_{j=1}^m \min_{i\in[1,n]} \|\mathbf{p}_j - f(\mathbf{x}_i)\|_2^2,
\end{equation}
\begin{equation}\label{eq:reg_2}
R_2(\mathbf{p}_1, ..., \mathbf{p}_m, D) = \frac{1}{n}\sum_{i=1}^n \min_{j\in[1,m]} \|f(\mathbf{x}_i) - \mathbf{p}_j\|_2^2.
\end{equation}
Here both terms are averages of minimum squared distances. The minimization of $R_1$ would require each prototype vector to be as close as possible to at least one of the training examples in the latent space. As long as we choose the decoder network to be a continuous function, we should expect two very close vectors in the latent space to be decoded to similar-looking images. Thus, $R_1$ will push the prototype vectors to have meaningful decodings in the pixel space. The minimization of $R_2$ would require every encoded training example to be as close as possible to one of the prototype vectors. This means that $R_2$ will cluster the training examples around prototypes in the latent space. We notice here that although $R_1$ and $R_2$ involve a minimization function that is not differentiable everywhere, these terms are differentiable almost everywhere and many modern deep learning libraries support this type of differentiation. Ideally, $R_{1}$ would take the minimum distance over the entire training set for every prototype; 
therefore, the gradient computation would grow linearly with the size of the training set. However, this would be impractical during optimization for a large dataset. To address this problem, we relax the minimization to be over only the random minibatch used by the Stochastic Gradient Descent (SGD) algorithm. For the other three terms, since each of them is a summation over the entire training set, it is natural to apply SGD to randomly selected batches for gradient computation.

Putting everything together, the cost function, denoted by $L$, on the training data $D$ with which we train our network $(f, g, h)$, is given by
\begin{align}
L((f, g, h), D) &= E(h \circ f, D) + \lambda R(g \circ f, D) \nonumber\\
&\quad+ \lambda_1 R_1(\mathbf{p}_1, ..., \mathbf{p}_m, D) \nonumber\\
&\quad+ \lambda_2 R_2(\mathbf{p}_1, ..., \mathbf{p}_m, D), \label{eq:cost}
\end{align}
where $\lambda$, $\lambda_1$, and $\lambda_2$ are real-valued hyperparameters that adjust the ratios between the terms.
 
\section{Case Study 1: Handwritten Digits}

We now begin a detailed walkthrough of applying our model to the well-known MNIST dataset. The Modified NIST Set (MNIST) is a benchmark dataset of gray-scale images of segmented and centered handwritten digits \cite{mnist}. We used 55,000 training examples, 5,000 validation examples, and 10,000 testing examples, where every image is of size $28 \times 28$ pixels. We preprocess the images so that every pixel value is in $[0, 1]$. This section is organized as follows: we first introduce the architecture and the training details, then compare the performance of our network model with other noninterpretible network models (including a regular convolutional neural network), and finally visualize the learned prototypes, the weight matrix $W$, and how a specific image is classfied.

\subsection{Architecture Details}
\citeauthor{autoencoder} \shortcite{autoencoder} show that a multilayer fully connected autoencoder network can achieve good reconstruction on MNIST even when using a very low dimensional latent space. We choose a multilayer convolutional autoencoder with a symmetric architecture for the encoder and decoder to be our model's autoencoder; these types of networks tend to reduce spatial feature extraction redundancy on image data sets and learn useful hierarchical features for producing state-of-the-art classification results. Each convolutional layer consists of a convolution operation followed by a pointwise nonlinearity. We achieve down-sampling in the encoder through strided convolution, and use strided deconvolution in the corresponding layer of the decoder.
After passing the original image through the encoder, the network flattens the resulted feature maps into a code vector and feeds it into the prototype layer. The resulting unflattened feature maps are fed into the decoder to reconstruct the original image. To visualize a prototype vector in the pixel space, we first reshape the vector to be in the same shape as the encoder output and then feed the shaped vector (now a series of feature maps) into the decoder.

The autoencoder in our network has four convolutional layers in both the encoder and decoder. All four convolutional layers in the encoder use kernels of size $3\times3$, same zero padding, and stride of size 2 in the convolution stage. The filters in the corresponding layers in the encoder and decoder are not constrained to be transposes of each other. Each of the outputs of the first three layers has 32 feature maps, while the last layer has 10. Given an input image of dimension $28\times28\times1$, the shape of the encoder layers are thus: $14\times14\times32$; $7\times7\times32$; $4\times4\times32$; $2\times2\times10$, and therefore the network compresses every 784-dimensional image input to a 40-dimensional code vector (2$\times$2$\times$10). Every layer uses the sigmoid function $\sigma(x)=\frac{1}{1+e^{-x}}$ as the nonlinear transformation. We specifically use the sigmoid function in the last encoder layer so that the output of the encoder is restricted to the unit hypercube $(0, 1)^{40}$. This allows us to initialize 15 prototype vectors uniformly at random in that hypercube. We do not use the rectified linear unit (ReLU -- \citeauthor{AlexNet}, \citeyear{AlexNet}) in the last encoder layer because using it would make it more difficult to initialize the prototype vectors, as initial states throughout ${\mathbb{R}_{\geqslant 0}}^{40}$ would need to be explored, and the network would take longer to stabilize. We also specifically choose the sigmoid function for the last decoder layer to make the range of pixel values in the reconstructed output $(0,1)$, roughly the same as the preprocessed image's pixel range.

\subsection{Training Details}
We set all the hyperparameters $\lambda$, $\lambda_{1}$, $\lambda_{2}$ to 0.05 and the learning rate to 0.0001. We minimize (\ref{eq:cost}) as a whole: we do not employ a greedy layer-wise optimization for different layers of the autoencoder nor do we first train the autoencoder and then the prototype classification network. 

Our goal in this work is not just to obtain  reasonable accuracy, but also interpretability. We use only a few of the general techniques for improving performance in neural networks, and it is possible that using more techniques would improve accuracy.
In particular, we use the data augmentation technique \textit{elastic deformation} \cite{Simard:2003} to improve prediction accuracy and reduce potential overfitting. The set of all elastic deformations is a superset of affine transformations. For every mini-batch of size 250 that we randomly sampled from the training set, we apply a random elastic distortion where a Gaussian filter of standard deviation equal to $4$ and a scaling factor of $20$ are used for the displacement field. Due to the randomness in the data augmentation process, the network sees a slightly different set of images during every epoch, which significantly reduces overfitting.

\subsection{Accuracy}
After training for 1500 epochs, our model achieved a classification accuracy of 99.53\% on the standard MNIST training set and 99.22\% on the standard MNIST test set. 

To examine how the two key elements of our interpretable network (the autoencoder and prototype layer) affect predictive power, we performed a type of ablation study. In particular, we trained two classification networks that are similar to ours, but removed some key pieces in both of the networks. The first network substitutes the prototype layer with a fully-connected layer whose output is a 15-dimensional vector, the same dimension as the output from the prototype layer; the second network also removes the decoder and changes the nonlinearity to ReLU. The second network is just a regular convolutional neural network that has similar architectural complexity to LeNet 5 \cite{mnist}. After training both networks using elastic deformation for 1500 epochs, we obtained test accuracies of 99.24\% and 99.23\% respectively. These test accuracies, along with the test accuracy of 99.2\% reported by \citeauthor{mnist} \shortcite{mnist}, are comparable to the test accuracy of 99.22\% obtained using our interpretable network. This result demonstrates that changing from a traditional convolutional neural network to our interpretable network architecture does not hinder the predictive ability of the network (at least not in this case).

In general, it is not always true that accuracy needs to be sacrificed to obtain interpretability; there could be many models that are almost equally accurate. The extra terms in the cost function (and changes in architecture) encourage the model to be more interpretable among the set of approximately equally accurate models.


\subsection{Visualization}
\begin{figure}[h!]
	\large
	\centering
	\includegraphics[width=0.8\columnwidth]{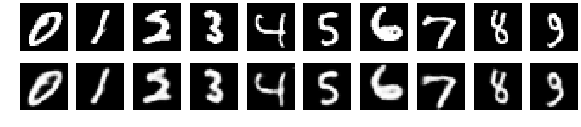}
	\caption{ Some random images from the training set in the first row and their corresponding reconstructions in the second row.}
	\label{fig:reconstruction}
\end{figure}

Let us first discuss the quality of the autoencoder, because good performance of the autoencoder will allow us to interpret the prototypes. After training, our network's autoencoder achieved an average squared $L^2$ reconstruction error of 4.22 over the undeformed training set, where examples are shown in Figure \ref{fig:reconstruction}. This reconstruction result assures us that the decoder can faithfully map the prototype vectors to the pixel space.
\begin{figure}[h!]
	\centering
	\includegraphics[width=0.9\columnwidth]{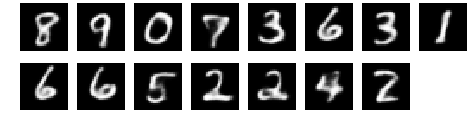}
	\caption{\footnotesize 15 learned MNIST prototypes visualized in pixel space.
	\label{fig:prototype}}
\end{figure}

We visualize the learned prototype vectors in Figure \ref{fig:prototype}, by sending them through the decoder. The decoded prototype images are sharp-looking and mostly resemble real-life handwritten digits, owing to the interpretability terms $R_1$ and $R_2$ in the cost function. Note that there is not a one-to-one correspondence between classes and prototypes.
Since we multiply the output of the prototype layer by a learnable weight matrix prior to feeding it into the softmax layer, the distances from an encoded image to each prototype have differing effects on the predicted class probabilities.

We now look at the transposed weight matrix connecting the prototype layer to the softmax layer, shown in Table \ref{weight_mnist},
\begin{table*}[t]
\centering
\begin{tabular}{r|r|r|r|r|r|r|r|r|r|r|}
\cline{2-11}
                                                                                    & 0                              & 1                              & 2                              & 3                              & 4                              & 5                              & 6                              & 7                              & 8                              & 9                              \\ \hline
\multicolumn{1}{|l|}{\raisebox{-0.1cm}{\includegraphics[height=\imsize]{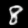}}}  & -0.07                          & 7.77                           & 1.81                           & 0.66                           & 4.01                           & 2.08                           & 3.11                           & 4.10                           & \cellcolor[gray]{0.75}-20.45 & -2.34                          \\ \hline
\multicolumn{1}{|l|}{\raisebox{-0.1cm}{\includegraphics[height=\imsize]{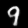}}}  & 2.84                           & 3.29                           & 1.16                           & 1.80                           & -1.05                          & 4.36                           & 4.40                           & -0.71                          & 0.97                           & \cellcolor[gray]{0.75}-18.10 \\ \hline
\multicolumn{1}{|l|}{\raisebox{-0.1cm}{\includegraphics[height=\imsize]{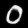}}}  & \cellcolor[gray]{0.75}-25.66 & 4.32                           & -0.23                          & 6.16                           & 1.60                           & 0.94                           & 1.82                           & 1.56                           & 3.98                           & -1.77                          \\ \hline
\multicolumn{1}{|l|}{\raisebox{-0.1cm}{\includegraphics[height=\imsize]{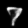}}}  & -1.22                          & 1.64                           & 3.64                           & 4.04                           & 0.82                           & 0.16                           & 2.44                           & \cellcolor[gray]{0.75}-22.36 & 4.04                           & 1.78                           \\ \hline
\multicolumn{1}{|l|}{\raisebox{-0.1cm}{\includegraphics[height=\imsize]{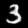}}}  & 2.72                           & -0.27                          & -0.49                          & \cellcolor[gray]{0.75}-12.00 & 2.25                           & -3.14                          & 2.49                           & 3.96                           & 5.72                           & -1.62                          \\ \hline
\multicolumn{1}{|l|}{\raisebox{-0.1cm}{\includegraphics[height=\imsize]{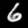}}}  & -5.52                          & 1.42                           & 2.36                           & 1.48                           & 0.16                           & 0.43                           & \cellcolor[gray]{0.75}-11.12 & 2.41                           & 1.43                           & 1.25                           \\ \hline
\multicolumn{1}{|l|}{\raisebox{-0.1cm}{\includegraphics[height=\imsize]{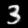}}}  & 4.77                           & 2.02                           & 2.21                           & \cellcolor[gray]{0.75}-13.64 & 3.52                           & -1.32                          & 3.01                           & 0.18                           & -0.56                          & -1.49                          \\ \hline
\multicolumn{1}{|l|}{\raisebox{-0.1cm}{\includegraphics[height=\imsize]{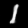}}}  & 0.52                           & \cellcolor[gray]{0.75}-24.16 & 2.15                           & 2.63                           & -0.09                          & 2.25                           & 0.71                           & 0.59                           & 3.06                           & 2.00                           \\ \hline
\multicolumn{1}{|l|}{\raisebox{-0.1cm}{\includegraphics[height=\imsize]{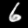}}}  & 0.56                           & -1.28                          & 1.83                           & -0.53                          & -0.98                          & -0.97                          & \cellcolor[gray]{0.75}-10.56 & 4.27                           & 1.35                           & 4.04                           \\ \hline
\multicolumn{1}{|l|}{\raisebox{-0.1cm}{\includegraphics[height=\imsize]{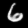}}}  & -0.18                          & 1.68                           & 0.88                           & 2.60                           & -0.11                          & -3.29                          & \cellcolor[gray]{0.75}-11.20 & 2.76                           & 0.52                           & 0.75                           \\ \hline
\multicolumn{1}{|l|}{\raisebox{-0.1cm}{\includegraphics[height=\imsize]{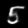}}} & 5.98                           & 0.64                           & 4.77                           & -1.43                          & 3.13                           & \cellcolor[gray]{0.75}-17.53 & 1.17                           & 1.08                           & -2.27                          & 0.78                           \\ \hline
\multicolumn{1}{|l|}{\raisebox{-0.1cm}{\includegraphics[height=\imsize]{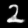}}} & 1.53                           & -5.63                          & \cellcolor[gray]{0.75}-8.78  & 0.10                           & 1.56                           & 3.08                           & 0.43                           & -0.36                          & 1.69                           & 3.49                           \\ \hline
\multicolumn{1}{|l|}{\raisebox{-0.1cm}{\includegraphics[height=\imsize]{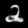}}} & 1.71                           & 1.49                           & \cellcolor[gray]{0.75}-13.31 & -0.69                          & -0.38                          & 4.55                           & 1.72                           & 1.59                           & 3.18                           & 2.19                           \\ \hline
\multicolumn{1}{|l|}{\raisebox{-0.1cm}{\includegraphics[height=\imsize]{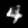}}} & 5.06                           & -0.03                          & 0.96                           & 4.35                           & \cellcolor[gray]{0.75}-21.75 & 4.25                           & 1.42                           & -1.27                          & 1.64                           & 0.78                           \\ \hline
\multicolumn{1}{|l|}{\raisebox{-0.1cm}{\includegraphics[height=\imsize]{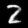}}} & -1.31                          & -0.62                          & -2.69                          & 0.96                           & 2.36                           & 2.83                           & 2.76                           & \cellcolor[gray]{0.75}-4.82  & -4.14                          & 4.95                           \\ \hline
\end{tabular}
\caption{Transposed weight matrix (every entry rounded off to 2 decimal places) between the prototype layer and the softmax layer. Each row represents a prototype node whose decoded image is shown in the first column. Each column represents a digit class. The most negative weight is shaded for each prototype. In general, for each prototype, its most negative weight is towards its visual class except for the prototype in the last row.
\label{weight_mnist}}
\end{table*}
to see the influence of the distance to each prototype on every class. We observe that each decoded prototype is visually similar to an image of a class for which the corresponding entry in the weight matrix has a significantly negative value. We will call the class to which a decoded prototype is visually similar the {\it visual class} of the prototype.


The reason for such a significantly negative value can be understood as follows. The prototype layer is computing the dissimilarity between an input image and a prototype through the squared $L^{2}$ distance between their representations in the latent space. Given an image $\mathbf{x}_i$ and a prototype $\mathbf{p}_j$, if $\mathbf{x}_i$ does not belong to the visual class of $\mathbf{p}_j$, then the distance between $f(\mathbf{x}_i)$ and $\mathbf{p}_j$ will be large, so that when $\|\mathbf{p}_j - f(\mathbf{x}_i)\|_2^2$ is multiplied by the highly negative weight connection between the prototype $\mathbf{p}_j$ and its visual class, the product will also be highly negative and will therefore significantly reduce the activation of the visual class of $\mathbf{p}_j$. As a result, the image $\mathbf{x}_i$ will likely not be classified into the visual class of $\mathbf{p}_j$. Conversely, if $\mathbf{x}_i$ belongs to the visual class of $\mathbf{p}_j$, then when the small squared distance $\|\mathbf{p}_j - f(\mathbf{x}_i)\|_2^2$ is multiplied by the highly negative weight connection between $\mathbf{p}_j$ and its visual class, the product will not decrease the activation of $\mathbf{p}_j$'s visual class too much. In the end, the activations of every class that $\mathbf{x}_i$ does not belong to will be significantly reduced because of some non-similar prototype, leaving only the activation of $\mathbf{x}_i$'s actual class comparatively large. Therefore, $\mathbf{x}_i$ is correctly classified in general.


An interesting prototype learned by the network is the last prototype in Table \ref{weight_mnist}. It is visually similar to an image of class 2; however, it has strong negative weight connections with class 7 and class 8 as well. Therefore, we can think of this prototype as being shared by these three classes, which means that an encoded input image that is far away from this prototype in latent space would be unlikely to be an image of 7, 8, or 2. This should not be too surprising: if we look at this decoded prototype image carefully, we can see that if we hide the tail of the digit, it would look like an image of 7; if we connect the upper-left endpoint with the lower-right endpoint, it would look like an image of 8.


Let us now look at the learned prototypes in Figure \ref{fig:prototype}. The three prototypes for class 6 seem to represent different writing habits in terms of what the loop and angle of ``6'' looks like. The first and third 6's have their loops end at the bottom while the second 6's loop ends more on the side. The 2's show similar variation. As for the two 3's, the two prototypes correspond to different curvatures.

Let us look into the model as it produces a prediction for a specific image of digit 6, shown on the left of Table \ref{distances}.
The distances computed by the prototype layer between the encoded input image and each of the prototypes are shown below the decoded prototypes in Table \ref{distances}, 
and the three smallest distances correspond to the three prototypes that resemble 6 after decoding. We observe here that these three distances are quite different, and the encoded input image is significantly closer to the third ``6'' prototype than the other two. This indicates that our model is indeed capturing the subtle differences within the same class.

After the prototype layer computes the 15-dimensional vector of distances shown in Table \ref{distances}, it is multiplied by the weight matrix in Table \ref{weight_mnist}, and the output is the unnormalized probability vector used as the logit for the softmax layer. The predicted probability of class 6 for this specific image is 99.99\%. 

\begin{table}[h!]
	\centering
	\begin{tabular}{llllll}
		&  \includegraphics[height=0.6cm]{0.png}& \includegraphics[height=0.6cm]{1.png} &  \includegraphics[height=0.6cm]{2.png}&  \includegraphics[height=0.6cm]{3.png}&  \includegraphics[height=0.6cm]{4.png}\\
		& 0.98 & 1.47 & 0.70 & 1.55 & 1.49 \\
		\includegraphics[height=0.6cm]{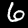}&  \includegraphics[height=0.6cm]{5.png}&  \includegraphics[height=0.6cm]{6.png}& \includegraphics[height=0.6cm]{7.png}&  \includegraphics[height=0.6cm]{8.png}&  \includegraphics[height=0.6cm]{9.png}\\
		& 0.29 & 1.69 & 1.02 & 0.41 & 0.15 \\
		& \includegraphics[height=0.6cm]{10.png}& \includegraphics[height=0.6cm]{11.png}&  \includegraphics[height=0.6cm]{12.png}&  \includegraphics[height=0.6cm]{13.png}&  \includegraphics[height=0.6cm]{14.png}\\
		& 0.88 & 1.40 & 1.45 & 1.28 & 1.28
	\end{tabular}
	\caption{The (rounded) distances between a test image 6 and every prototype in the latent space.}
	\label{distances}
\end{table}

\section{Case Study 2: Cars}

\begin{figure}[h]
	\centering
	\includegraphics[width=\columnwidth]{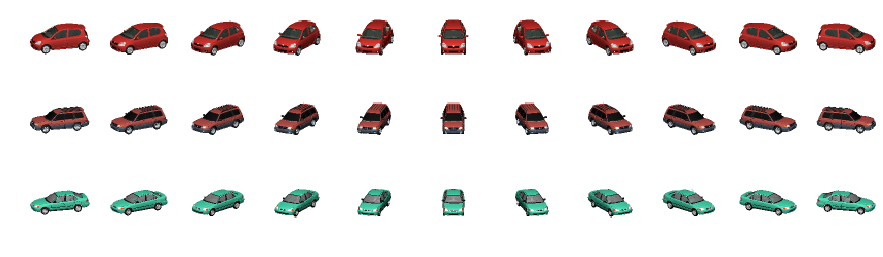}
	\caption{Three cars at 11 angles from car dataset.}
	\label{fig:Car_dataset}
\end{figure}

The second dataset we use consists of rendered color images, each with 64 $\times$ 64 $\times$ 3 pixels, of 3D car models with varying azimuth angles at $15^{\circ}$ intervals, from $-75^{\circ}$ to $75^{\circ}$ \cite{car_dataset}. There are 11 views of each car and every car's class label is one of the 11 angles (see Figure \ref{fig:Car_dataset}). The dataset is split into a training set ($169\times11=1859$ images) and a test set ($14\times11 =154$ images).

\begin{figure}[h!]
	\centering
	\includegraphics[width=0.8\columnwidth]{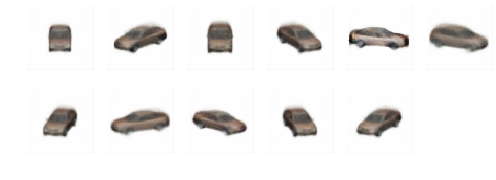}
	\caption{Decoded prototypes when we include $R_1$ and $R_2$.
	\label{fig:both_R1_and_R2}}
\end{figure}

We use two convolutional layers in both the encoder and decoder. The first and the second layer in the encoder uses respectively 32 and 10 convolutional filters of size 5 $\times$ 5, stride 2, and no zero padding. The architecture of the decoder is symmetric to that of the encoder. We use the sigmoid activation function in the last layer of the encoder and the decoder, and leaky ReLU in all other autoencoder layers. We set the number of our prototypes to be eleven, which is the same as the number of classes. Figure \ref{fig:both_R1_and_R2} shows the eleven decoded prototypes from our model. If we compare Figure \ref{fig:Car_dataset} and Figure \ref{fig:both_R1_and_R2} in color, we can observe that the network has determined that the color of a car is not important in determining the angle, so all of the decoded prototypes are of the same ``average'' color. The learned weight matrix $W$ is shown in Table 4 in the Supplementary Material. We compared our model to a network without the interpretable parts, in which we removed the decoder and replaced the prototype layer with a fully connected layer of the same size. The accuracies for these two models are shown in Table \ref{car_accuracy}. The result again illustrates that we do not sacrifice much accuracy when including the interpretability elements into the network.

\begin{table}[h!]
\centering
\begin{tabular}{ c | c | c }
		\hline
		 & interpretable  &  non-interpretable  \\ \hline
		train acc &98.2\%  &99.8\%  \\ \hline		
		test acc &93.5\%  & 94.2\% \\ \hline		
	\end{tabular}
	\caption{Car dataset accuracy.
	\label{car_accuracy}}
\end{table} 

\begin{figure}[h!]
	\centering
	\includegraphics[width=0.8\columnwidth]{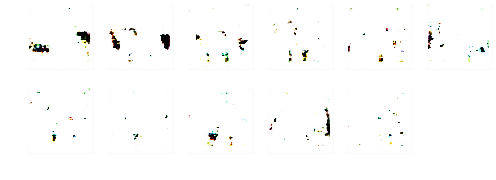}
	\caption{Decoded prototypes when we remove $R_1$ and $R_2$.
	\label{fig:Remove_R1_and_R2}}
\end{figure}

\begin{figure}[h!]
	\centering
	\includegraphics[width=0.8\columnwidth]{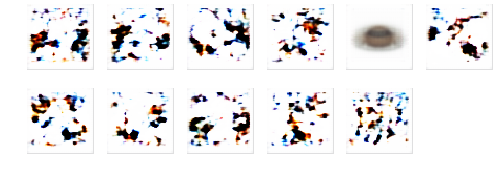}
	\caption{Decoded prototypes when we remove $R_1$.
	\label{fig:Remove_R1}}
\end{figure}

\begin{figure}[h!]
	\centering
	\includegraphics[width=0.8\columnwidth]{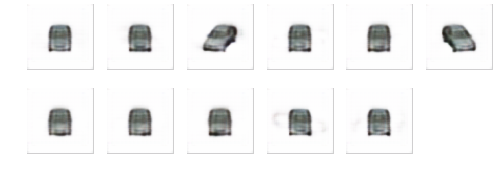}
	\caption{Decoded prototypes when we remove $R_2$.
	\label{fig:Remove_R2}}
\end{figure}

We use this case study to illustrate the importance of the two interpretability terms $R_1$ and $R_2$ in our cost function. If we remove both $R_1$ and $R_2$, the decoded prototypes will not look like real images, as shown in Figure \ref{fig:Remove_R1_and_R2}. 
If we leave out only $R_1$, the decoded prototypes will again not look like real observations, as shown in Figure \ref{fig:Remove_R1}. If we remove only $R_2$, the network chooses prototypes that do not fully represent the input space, and some of the prototypes tend to be similar to each other, as shown in Figure \ref{fig:Remove_R2}. Intuitively, $R_1$ pushes every prototype to be close to a training example in the latent space so that the decoded prototypes can be realistic, while $R_2$ forces every training example to find a close prototype in the latent space, thereby encouraging the prototypes to spread out over the entire latent space and to be distinct from each other.  
In other words, $R_1$ helps make the prototypes meaningful, and $R_2$ keeps the explanations faithful in forcing the network to use nearby prototypes for classification.

\section{Case Study 3: Fashion MNIST} Fashion MNIST \cite{fashion_mnist} is a dataset of Zalando's article images, consisting of a training set of 60,000 examples and a test set of 10,000 examples. Each example is a $28\times28$ grayscale image, associated with a label from 10 classes, each being a type of clothes item. The dataset shares the same image size and structure of training and testing splits as MNIST.

\begin{figure}[h]
	\centering
	\includegraphics[width=0.8\columnwidth]{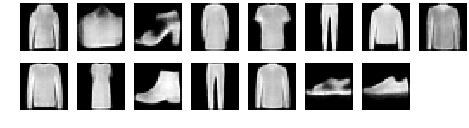}
	\caption{15 decoded prototypes for Fashion-MNIST.
	\label{fig:fmnist}}
\end{figure}

We ran the same model from Case Study 1 on this fashion dataset and achieved a testing accuracy of 89.95\%. This result is comparable to those obtained using standard convolutional neural networks with max pooling reported on the dataset website (87.6-92.5\% for networks that use similar architecture complexity as ours, \citeauthor{FashionMNISTWebsite}, \citeyear{FashionMNISTWebsite}). The learned prototypes are shown in Figure \ref{fig:fmnist}. For each class, there is at least one prototype representing that class. The learned prototypes have fewer details (such as stripes, precence of a collar, texture) than the original images. This again shows that the model has recognized what information is important in this classification task -- the contour shape of the input is more useful than its fine-grained details. The learned weight matrix $W$ is shown in Table 5 in the Supplementary Material.

\section{Discussion and Conclusion}
We combine the strength of deep learning and the interpretability of case-based reasoning to make an interpretable deep neural network. The prototypes can provide useful insight into the inner workings of the network, the relationship between classes, and the important aspects of the latent space, as demonstrated here. Although our model does not provide a full solution to problems with accountability and transparency of black box decisions, it does allow us to partially trace the path of classification for a new observation.

We have noticed in our experiments that the addition of the two interpretability terms $R_1$ and $R_2$ tend to act as regularizers and help to make the network robust to overfitting. The extent to which interpretability reduces overfitting is a topic that could be explored in future work.

\noindent\textbf{Supplementary Material and Code:} Our supplementary material and code are available at this URL: \url{https://github.com/OscarcarLi/PrototypeDL}.

\noindent\textbf{Acknowledgments:} This work was sponsored in part by MIT Lincoln Laboratory.

\fontsize{9.5pt}{10.5pt}
\selectfont
\bibliographystyle{aaai}
\bibliography{li_liu_chen_rudin}
\end{document}